%% file: acl_latex.tex
\pdfoutput=1

\documentclass[11pt]{article}

\usepackage[final]{acl}

\usepackage{times}
\usepackage{latexsym}

\usepackage[T1]{fontenc}

\usepackage[utf8]{inputenc}

\usepackage{microtype}

\usepackage{inconsolata}

\usepackage{graphicx}

\usepackage{hyperref}
\usepackage{xurl}
\usepackage{tabularx}
\usepackage{multirow}
\usepackage{booktabs}
\usepackage{graphicx}
\usepackage{xspace}
\usepackage{subcaption}
\usepackage{changepage}

\newcommand{\method}{ACID\xspace}

\newcommand{\ensuretext}[1]{#1}
\newcommand{\marker}[2]{\ensuremath{^{\textsc{#1}}_{\textsc{#2}}}}
\newcommand{\arkcomment}[3]{\ensuretext{\textcolor{#3}{[#1 #2]}}}

\newcommand{\nascomment}[1]{\arkcomment{\marker{NA}{S}}{#1}{purple}}
\newcommand{\jungo}[1]{\arkcomment{\marker{J}{K}}{#1}{blue}}

\title{Summarization-Based Document IDs for Generative Retrieval with Language Models}

\author{Alan Li$^1$ \quad Daniel Cheng$^1$ \quad Phillip Keung$^3$ \quad Jungo Kasai$^1$ \quad Noah A. Smith$^{1,2}$ \\
  $^1$Paul G. Allen School of Computer Science \& Engineering, University of Washington, USA \\
  $^2$Allen Institute for Artificial Intelligence, USA \\
  $^3$Department of Statistics, University of Washington, USA \\
  \texttt{\{lihaoxin,d0,jkasai,nasmith\}@cs.washington.edu}, 
  \texttt{pkeung@uw.edu}} 

\begin{document}
\maketitle
\begin{abstract}
Generative retrieval \citep{pawa, neural_indexing} is a popular approach for end-to-end document retrieval that directly generates document identifiers given an input query. We introduce \emph{summarization-based document IDs}, in which each document's ID is composed of an extractive summary or abstractive keyphrases generated by a language model, rather than an integer ID sequence or bags of n-grams as proposed in past work. We find that abstractive, content-based IDs (\method) and an ID based on the first 30 tokens are very effective in direct comparisons with previous approaches to ID creation. We show that using \method improves top-10 and top-20 recall by 15.6\% and 14.4\% (relative) respectively versus the cluster-based integer ID baseline on the MSMARCO 100k retrieval task, and 9.8\% and 9.9\% respectively on the Wikipedia-based NQ 100k retrieval task. Our results demonstrate the effectiveness of human-readable, natural-language IDs created through summarization for generative retrieval. We also observed that extractive IDs outperformed abstractive IDs on Wikipedia articles in NQ but not the snippets in MSMARCO, which suggests that document characteristics affect generative retrieval performance. 
\end{abstract}


\input{sections/intro}

\input{sections/methods}
\input{sections/experiments}
\input{sections/results}
\input{sections/related}
\input{sections/discussion}

\section*{Limitations}

Due to constraints on our computational budget, the largest dataset that we used contains 100k query-document pairs, which is a subset of the full NQ or MSMARCO datasets, and the largest model that we trained was the 2.8-billion parameter Pythia model, which is not the largest model in the Pythia model family. We expect that the performance characteristics of our method may change as the datasets and models are scaled up to sizes that practitioners in industry settings would typically use.

\bibliography{acl_latex}

\end{document}

%% file: sections/intro.tex
\section{Introduction}

Wikipedia-based corpora have long been an important part of NLP research and form a natural benchmark for studying new techniques in text-based recommender and information retrieval systems. In this work, we examine how \emph{generative retrieval} behaves on short-form and long-form documents drawn from Wikipedia and non-Wikipedia sources. We also propose a new type of document ID for generative retrieval based on \emph{document summarization}, which demonstrably improves retrieval performance across the tasks that we examined.

Large language models (LMs) are now widely used across many NLP tasks, and extensions of generative models to document retrieval tasks have recently been proposed \citep{pawa, neural_indexing}, in contrast to vector-based approaches like dense passage retrieval (DPR; \citealp{dpr}). DPR is a widely-used technique for training document retrieval models, where queries and documents are mapped to dense vector representations with a transformer encoder (e.g., BERT; \citealp{bert}). By increasing the cosine similarity between positive query-document pairs and decreasing it between negative pairs, DPR performs metric learning over the space of queries and the set of documents to be indexed. 




Generative alternatives to document retrieval address certain limitations of dense, vector-based approaches to retrieval. For example, query and document representations are constructed separately in DPR, which precludes complex query-document interactions. Using a single dense vector to represent an entire document limits the amount of information that can be stored; indeed, \citet{neural_indexing} observed that increasing the number of parameters in the encoder does not significantly enhance DPR performance. Furthermore, the rich sequence generation capabilities of language models (LMs) cannot be used directly in dense retrieval. \citet{neural_indexing} and \citet{pawa} therefore proposed a new direction called \emph{generative retrieval}, where LMs learn to directly map queries to an identifier that is unique to each document. We illustrate the differences in Figure \ref{fig:dpr_vs_gen}.

\input{figures/dpr_vs_gen}

Instead of retrieving documents based on cosine similarity, generative retrieval uses an LM to produce a sequence of tokens encoding the relevant document's ID, conditional on the query. Decoding constraints are applied to ensure that only document IDs that exist in the corpus are generated. \citet{neural_indexing} and \citet{pawa} showed that generative retrieval outperformed DPR on information retrieval benchmarks like Natural Questions \citep{nq} and TriviaQA \citep{trivia_qa}, and subsequent publications have corroborated their findings on other retrieval tasks like multilingual retrieval \citep{dsi_qg}.


State-of-the-art generative retrieval models rely on document clustering to create document IDs, following the work of both \citet{pawa} and \citet{neural_indexing}, and the resulting document ID is an integer sequence corresponding to the clusters that the document belongs to. However, generating arbitrary sequences of integers is very different from what LMs are designed to do, since LMs are pretrained to generate natural language. In addition to negatively impacting LM generation performance, cluster-based integer IDs are not human-readable and require re-clustering if a substantial number of new documents are added to the index. 

To address the issues with cluster-based IDs, we consider \emph{summarization-based document IDs}, which are human-readable, natural-language document IDs. We propose \textbf{\method}, an \textbf{A}bstractive, \textbf{C}ontent-based \textbf{ID} assignment method for documents, alongside simpler IDs based on extractive summarization. \method uses a language model (GPT-3.5 in our experiments) to generate a short sequence of \emph{abstractive keyphrases} from the document's contents to serve as the document ID, rather than a hierarchical clustering ID or an arbitrary integer sequence. 
We also consider creating content-based IDs extractively:   taking the first 30 tokens of each document as its ID or choosing the top-30 keywords with respect to BM25 scores.
We find that \method generally outperforms the cluster-based IDs for generative retrieval (as well as the extractive methods) in direct comparisons on standard retrieval benchmarks. We also observe that longer extractive document IDs are helpful for retrieving long documents, such as the Wikipedia articles in the NQ benchmark, versus the shorter document fragments from the MSMARCO dataset.



Finally, we examine the effect of hyperparameters like model size and beam width on retrieval performance, and compare how cluster-based IDs and summarization-based IDs behave under different settings. 

The code for reproducing our results and the keyword-augmented datasets can be found at \url{https://github.com/lihaoxin2020/Summarization-Based-Document-IDs-for-Generative-Retrieval}, and the data can be found at \url{https://huggingface.co/datasets/lihaoxin2020/abstractive-content-based-IDs}.

%% file: figures/dpr_vs_gen.tex
\begin{figure*}[h]
\centering
\includegraphics[width=0.99\linewidth]{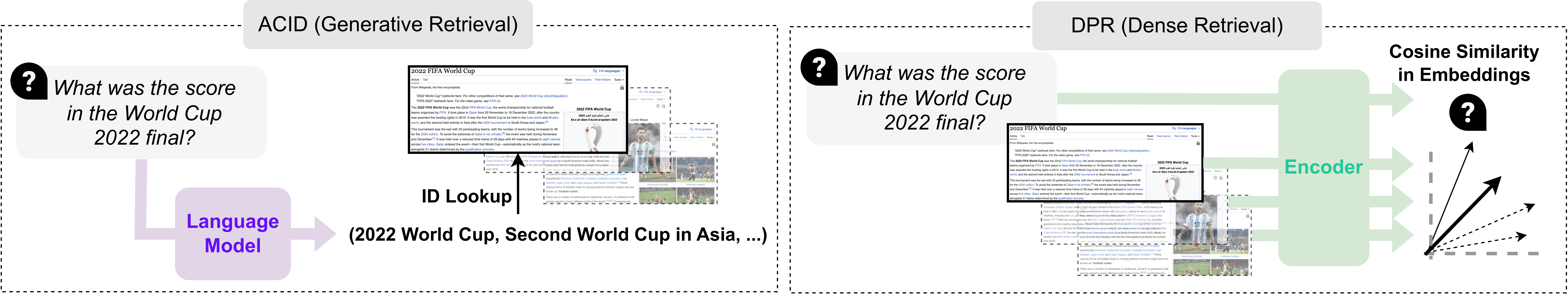}
\caption{
Generative retrieval vs.\ dense retrieval. 
In dense retrieval (right), both the query and the documents are encoded into \emph{dense} vectors (i.e., embeddings).
Nearest-neighbor search is then applied to find the most relevant documents.
Generative retrieval (left) trains a language model to generate the relevant document ID conditional on the query. The ID is tied to a unique document, allowing for direct lookup.
We propose summarization-based document IDs like \method, which uses GPT-3.5 to create a sequence of abstractive keyphrases to serve as the document ID. 
} 
\label{fig:dpr_vs_gen}
\end{figure*}

%% file: sections/methods.tex
\section{IDs for Generative Retrieval}

\input{tables/example}

Since generative retrieval is a comparatively new approach for document retrieval, there is significant variation in the literature on how language models are trained to map queries to document IDs. \citet{neural_indexing} distinguish between the `indexing' step (where the LM is trained to link spans from the training, development, and test documents to their document IDs) and the `finetuning' step (where the training query-document pairs are used to finetune the LM for retrieval). Note that generative retrieval models must index all documents, including the development and test documents, in order for the language model to be aware of their document IDs at inference time. Additionally, \citet{pawa} and \citet{dsi_qg} perform data augmentation in the indexing and finetuning steps by introducing `synthetic' queries, where a query generation model \citep{doc2query} based on T5 \citep{t5} generates additional queries for each document. 

In the three subsections that follow, we elaborate on each of the steps for generative retrieval. Figure \ref{fig:architecture} depicts the steps needed to create our summarization-based document IDs, perform data augmentation, index the documents with the LM, and finetune the LM for generative retrieval.

\subsection{Document ID Creation}

In Table \ref{table:sample}, we provide an example of a document about engineering sub-disciplines and the cluster-based and content-based IDs that would be derived from it. From the example, it is clear why we would expect \method to outperform cluster IDs, since it is straightforward for LMs to generate the keyphrase sequence given an engineering-related query. The cluster ID, on the other hand, resembles an integer hash of the document (with some semantic information carried over from the clustering).

\textbf{Abstractive, Content-based IDs.} We create natural language IDs for every document to be indexed by generating keyphrases. Tokens from the document (up to the maximum context size of 4000 tokens) are used as part of a prompt to an LM to generate 5 keyphrases. The keyphrases are a brief abstractive summary of the topics in the document. The keyphrases are concatenated together to form the \method for each document. We create IDs for every document in the training, development, and test sets. 


We chose the GPT-3.5 API provided by OpenAI to generate keyphrases, though any reasonable pretrained LM can be used instead. The prompt that we used was:
\begin{adjustwidth}{0.3cm}{0.3cm}
\begin{quote}
Generate no more than 5 key phrases describing the topics in this document. Do not include things like the Wikipedia terms and conditions, licenses, or references section in the list: (document body here)
\end{quote}
\end{adjustwidth}

\textbf{Extractive Summary IDs.} We consider two types of extractive summary IDs: a bag of unigrams selected based on BM25 scores, and the first $k$ tokens of the document. For many types of documents (e.g., news articles, Wikipedia articles, scientific papers), the first few sentences would generally provide an overview of the contents of the document, which motivates our choice of the first $k$ tokens as a kind of extractive document ID.

\textbf{Cluster-based IDs.} By way of comparison with our proposed IDs, cluster-based IDs are integer sequences. An encoder creates an embedding vector for each document in the dataset, and the document embeddings are clustered using the $k$-means algorithm. If the number of documents in a cluster exceeds a predefined maximum, then subclusters are created recursively, until all subclusters contain fewer documents than the maximum. Each document's ID is a sequence of integers, corresponding to the path to the document through the tree of hierarchical clusters. The number of clusters at each level and the maximum number of documents in each cluster are hyperparameters. (For example, the values reported by \citealp{pawa}, were 10 and 100 respectively, which we also use in our experiments.) 



\input{figures/architecture}

\subsection{Document Indexing and Supervised Finetuning}

We first index all of the documents in the training, development, and test sets. For indexing purposes, we consider input/output pairs of the form 

\begin{itemize}
    \item (synthetic query, document ID).
\end{itemize}

In other words, the LM is trained to generate the relevant document ID, given a randomly selected document span or a synthetic query, as part of the indexing task. We use a T5-based query generation model to provide synthetic queries given the body of each document, which serves as a form of data augmentation independent of the queries in the training data. Note that, in our experiments, only synthetic queries are used during the indexing step. Although random document spans are used in other generative retrieval papers, we did not observe an improvement by doing so. 



After document indexing, we finetune the model on the retrieval training data:
\begin{itemize}
    \item (user-generated query, document ID)
\end{itemize}
In other words, the LM is trained to generate the document ID, given a real, user-generated query.


\subsection{Retrieving Documents}

At inference time, the LM generates a document ID via beam search, given a user-generated query from the test set. We use a constrained decoder at inference time, which is constrained by a prefix tree such that it can only generate document IDs that exist in the corpus. Since each document ID maps to a unique document, it is straightforward to compute the proportion of queries for which the model retrieved the correct document. Model performance is measured based on the recall of relevant documents retrieved within the top-1, top-10, and top-20 results in our experiments.

%% file: tables/example.tex
\begin{table*}[h]
\centering
\small

\begin{tabularx}{\linewidth}{@{}X@{}}
\toprule
\textbf{Document Text} \\
\midrule
List of engineering branches Engineering is the discipline and profession that applies scientific theories , mathematical methods , and empirical evidence to design , create , and analyze technological solutions cognizant of safety , human factors , physical laws , regulations , practicality , and cost . In the contemporary era , engineering is generally considered to consist of the major primary branches of chemical engineering , civil engineering , electrical engineering , and mechanical engineering… \\
\midrule
\textbf{Cluster-based Document ID} \\
\midrule
9, 5, 1, 9, 6, 1, 0, 4, 8, 1, 3, 1, 2, 9, 0 \\
\midrule
\end{tabularx}
\begin{tabularx}{\linewidth}{@{}XXX@{}}
\multicolumn{3}{@{}l}{\textbf{Summarization-based Document IDs}} \\
\midrule
\emph{First $k$ Tokens} & \emph{BM25 Scoring} & \emph{\method} \\
\midrule
List of engineering branches Engineering is the discipline and profession that applies scientific theories , mathematical... & teletraffic optomechanical nanoengineering subdiscipline eegs biotechnical bioprocess mechatronics metallics crazing... & (1) Major engineering branches: chemical, civil, electrical, mechanical (2) Chemical engineering: conversion of raw materials with varied specialties (3) Civil engineering: design… \\
\bottomrule
\end{tabularx}
\caption{An example of a document, its cluster-based ID (where each level of the clustering has 10 clusters), and its associated natural language, content-based IDs. `First $k$ tokens' sets the ID to be the document's first $k$ tokens. BM25 scoring uses the top-$k$ highest-scoring tokens from the document as the ID, where scores are based on Okapi BM25. \method uses an LM (e.g., GPT-3.5) to generate 5 keyphrases as the ID.}
\label{table:sample}
\end{table*}

%% file: figures/architecture.tex
%
\begin{figure*}[h]
\centering
\includegraphics[width=0.9\linewidth]{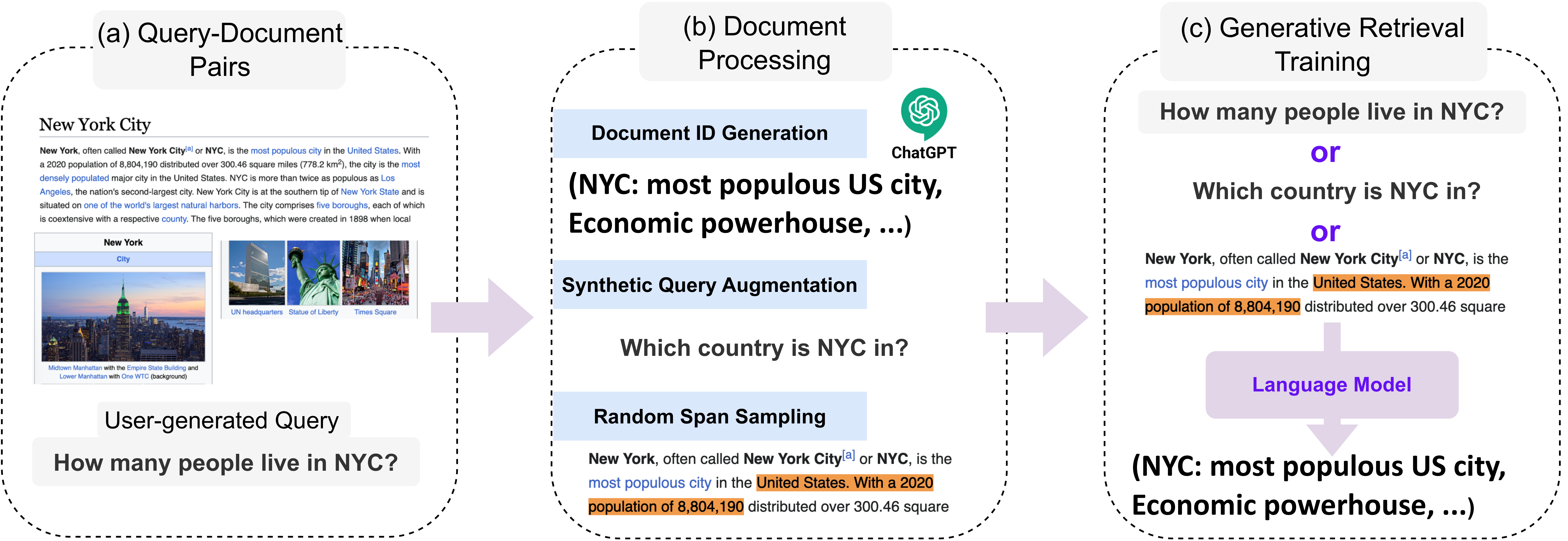}
\caption{Data processing and model training. (a) Each document-query pair from the training corpus will be converted into inputs and outputs for finetuning the pretrained transformer decoder, which serves as the generative retrieval model. (b) GPT-3.5 is used to generate a sequence of keyphrases, which is used as the document ID. (c) Given a user query or a synthetic query, the generative retrieval model learns to generate the ID of the relevant document. We use a doc2query model to generate synthetic queries as additional inputs.
Randomly sampled spans of 64 tokens can also be used as inputs to ensure that the model associates the contents of each document with its ID.
} 
\label{fig:architecture}
\end{figure*}

%% file: sections/experiments.tex
\section{Experiments}

In the experiments below, we demonstrate that summarization-based IDs outperform cluster-based IDs on the NQ and MSMARCO retrieval benchmarks. Simple extractive IDs, like using the first 30 tokens of the document or BM25-based keyword selection, can outperform the cluster-based approach in most cases. We also compare our IDs with another keyword-based document ID method that constructs IDs using learned relevance scores \citep{tsgen}. We then show that summarization-based IDs work well across a range of language model sizes (as measured by the total number of parameters). Finally, we show that widening the beam improves retrieval performance meaningfully for ACID, whereas cluster-based IDs benefit from beam width to a lesser degree (or not at all, in the case of the widest beam widths). 


The BM25-based IDs were created by ranking all of the unique terms in each document by their BM25 scores, and taking the top 30 terms as the document ID. We used Anserini \citep{anserini} to compute BM25 scores for the documents in each corpus. To avoid selecting very rare terms as part of each document's BM25-based document ID, we required that each term either appear at least 2 times in the document itself, or appear at least 5 times in the corpus. 

We use the Natural Questions (NQ; \citealp{nq}) and MSMARCO \citep{msmarco} datasets. For each dataset, we finetune a pretrained language model for retrieval on 1k, 10k, and 100k random samples of the training split. Note that MSMARCO and NQ do not disclose their test sets publicly, and our results are reported on the provided development sets. Since we did not use the entirety of the training data that was available for NQ and MSMARCO, we created separate development sets for them by taking a random sample of each dataset's training data. We provide the details of each corpus in Table \ref{tab:stats}. Document length is highly variable, and we truncate all documents after 4k tokens.

We use the Pythia LMs \citep{pythia} to initialize the retrieval model in our experiments. All of our models are trained on AWS g5 instances equipped with Nvidia A10G GPUs. Models are optimized using AdamW \citep{adamW}. We provide the model hyperparameters that were used in the Appendix. The beam width for all experiments is 20, unless stated otherwise.

In Table \ref{tab:stats}, we provide the basic statistics for the NQ and MSMARCO datasets that we used. We deduplicate documents based on the first 512 tokens of each document, and documents with $\geq$95\% token overlap are considered duplicates. 

Note that there is a substantial difference in the average document length between NQ and MSMARCO datasets. While NQ and MSMARCO have queries of similar lengths, their document lengths are very different, since NQ documents are complete Wikipedia articles while MSMARCO passages are a few sentences long, excerpted from a longer document.


\input{tables/models_and_data}

%% file: tables/models_and_data.tex
\begin{table}[h]
\small
\centering

\begin{tabular}{@{}lrrr@{}}
\toprule
&  & Ave. Query & Ave. Doc.  \\
& \# Pairs & Length & Length \\
\midrule
NQ-100k      & 100,000 & \multirow{3}{*}{49.2} & \multirow{3}{*}{36,379.4} \\
NQ-Dev       & 1,968 &  &  \\
NQ-Test      & 7,830 &  &  \\
\midrule
MSMARCO-100k & 100,000 & \multirow{3}{*}{32.8} & \multirow{3}{*}{334.4} \\
MSMARCO-Dev  & 2,000 &  &  \\
MSMARCO-Test & 6,980 &  &  \\
\bottomrule

\end{tabular}
\caption{Dataset characteristics. `\# Pairs' refers to the number of query-document pairs. Average lengths refer to the average length in characters.}
\label{tab:stats}
\end{table}

%% file: sections/results.tex
\section{Results}


There is substantial variation in the reported results on the NQ dataset among papers that use cluster-based IDs for generative retrieval. In \citet{neural_indexing} and \citet{pawa}, the top-1 recall with the NQ 320k dataset were 27.4\% and 65.86\% respectively, despite both groups using the same T5-Base model initialization and cluster-based ID approach. There are many possible explanations for the discrepancy (e.g., use of synthetic queries, computational budget, etc.), but at the time of writing, neither paper has made the code or processed data publicly available, which makes replication difficult. For this reason, we focus on internal comparisons rather than external ones, where we control the relevant experimental settings to ensure that the comparisons are fair and the differences in results are meaningful.

\subsection{MSMARCO}

\input{tables/msmarco}

We begin by examining the performance of our implementations of various types of document IDs on the MSMARCO task. We present the results in Table \ref{tab:msmarco}, and all results are based on a 160M-parameter pretrained Pythia LM. Across all training set sizes, the {\method}s offer better retrieval performance compared to the other ID generation techniques, and summarization-based IDs clearly outperform the cluster integer IDs.

\subsection{Natural Questions}
\label{sec:nq_results}
\input{tables/nq}

In Table \ref{tab:nq}, we compare sparse and dense retrieval techniques against generative retrieval on the NQ dataset. We used the 160M-parameter Pythia LM as our base model to obtain the results in the table. Across the NQ 1k, 10k, and 100k tasks, summarization-based document IDs generally outperform cluster-based integer IDs and TSGen \citep{tsgen}. (TSGen learns a scoring function that identifies relevant terms from the document to use as the ID.) As we saw with MSMARCO, the simple approach of using the first 30 tokens from each document to create IDs also outperforms the cluster-based approach.

We further improve the performance of the finetuned 160M-parameter model by performing joint decoding with the 12-billion parameter Pythia LM. We provide 8 query-document ID pairs from the training data to the 12B Pythia model for in-context learning. For a given query, we use both the small model and the large model (with the in-context examples) to generate the relevant document ID. The output probabilities from the small and large models are combined using a mixture weight of $\alpha=0.85$ on the small model.

When we applied joint decoding, the extractive summarization-based document ID that uses the first 30 tokens outperformed all of the other techniques that we examined. 


We emphasize that this is one of the major advantages of using generative retrieval with natural-language IDs: we can use a pretrained LLM with in-context learning to significantly boost the performance of a smaller finetuned LM. In contrast, generative retrieval that uses integer IDs does not benefit from joint decoding with an LLM, since the integer ID sequences are far from the pretraining distribution and in-context learning provides no benefit.

We observed that the top-1 recall with the first 30 tokens as the ID is quite high. This may be due to the structure of the NQ documents, which are Wikipedia articles. The first tokens of every document are the title of the Wikipedia page, and so the first 30 tokens represent a very effective ID for retrieval purposes. Nonetheless, without joint decoding, {\method} outperforms the first 30 token IDs at top-10 and top-20 recall. 


\subsection{Model Size}

\input{figures/model_size}

We examine whether the relative outperformance of {\method}s versus cluster integer IDs on MSMARCO is affected by the number of parameters in the generative model. Our default experiments in the previous sections used 160M-parameter Pythia models, and in Figure \ref{fig:model_size} we conduct experiments going up to 2.8B-parameter models.

We observe that {\method}s continue to outperform cluster integer IDs, even as we vary the model size. In general, increasing the size of the model leads to an improvement in retrieval performance, regardless of the ID type.


\subsection{Beam Width}

\input{figures/beams}

From Table \ref{fig:beams}, we see that larger beam widths generally improve recall on MSMARCO, though with rapidly diminishing returns. The top-1 recall does not benefit past a beam width of 8, and the recall rapidly plateaus as beam width increases from 1 to 16. This is true for both cluster integer IDs and \method, though \method does benefit more in absolute terms than cluster IDs from a wider beam (when comparing a beam width of 1 to a beam width of 16). 

In the same table, we also examine the effect of very wide beams on recall at 10 and 20 for the MSMARCO dataset. Some benefit is observed when \method is the document ID, but no improvement is observed for cluster IDs. 


As discussed previously, the cluster integer ID is typically restricted to a small number of clusters per level (the digits 0 through 9, for example), and so a wide beam in excess of that number doesn't yield any improvements, whereas \method does benefit from wider beams, since it is a natural-language ID with access to the full vocabulary of the LM.


\subsection{ID Length}


\input{tables/num_first_words}

In Table \ref{tab:num_first_words}, we present the change in recall on the NQ and MSMARCO tasks depending on the length of the document ID. We use the extractive document ID based on the first 10, 20, 30, and 40 tokens. On MSMARCO 100k, we observe very little change in top-k recall. On NQ 100k, we saw a larger benefit with longer IDs, with the highest recall corresponding to the longest document ID. We speculate that the differences in document length between MSMARCO and NQ ({$\sim\hspace{-0.3em}334$} tokens versus {$\sim\hspace{-0.3em}36$}k tokens per document) means that longer IDs tend to benefit the NQ retrieval task more. 




%% file: tables/msmarco.tex
\begin{table*}[h]
\centering
\small

\makebox[\textwidth][c]{
\begin{tabular}{@{}lccccccccc@{}}
\toprule
 & \multicolumn{3}{c}{MSMARCO 1k} & \multicolumn{3}{c}{MSMARCO 10k} & \multicolumn{3}{c}{MSMARCO 100k} \\
 & Rec@1 & @10 & @20 & Rec@1 & @10 & @20 & Rec@1 & @10 & @20 \\
\midrule
\emph{Baseline} &&&&&&&&&\\
\midrule
Cluster Integer IDs   & 41.1 & 59.5 & 64.2 & 42.4 & 62.3 & 67.1 & 46.8 & 68.8 & 73.4 \\
\midrule
\emph{Extractive Summarization IDs} &&&&&&&&&\\
\midrule
BM25 Top-30   & 48.7 & 74.3 & 79.4 & 49.1 & 75.7 & 80.1 & 52.0 & 79.2 & 82.9 \\
First 30 Tokens     & 49.0 & 73.0 & 77.8 & 48.7 & 72.8 & 77.9 & 51.8 & 76.0 & 79.6 \\
\midrule
\emph{Abstractive Summarization IDs} &&&&&&&&&\\
\midrule
{\method}       & \textbf{49.1} & \textbf{74.3} & \textbf{80.1} & \textbf{50.4} & \textbf{76.3} & \textbf{80.4} & \textbf{52.9} & \textbf{79.5} & \textbf{84.0} \\
\bottomrule
\end{tabular}}
\caption{Recall for MSMARCO. Recall refers to the percentage of queries in the evaluation set for which the ground-truth document ID was produced in the top-1, top-10, and top-20 candidates from constrained beam search decoding. MSMARCO 1k, 10k, and 100k refer to the number of training query-document pairs used to finetune the LM.}
\label{tab:msmarco}
\end{table*}

%% file: tables/nq.tex
\begin{table*}[h]
\small
\centering

\makebox[\textwidth][c]{
\begin{tabular}{@{}lccccccccc@{}}
\toprule
 & \multicolumn{3}{c}{NQ 1k} & \multicolumn{3}{c}{NQ 10k} & \multicolumn{3}{c}{NQ 100k} \\
 & Rec@1 & @10 & @20 & Rec@1 & @10 & @20 & Rec@1 & @10 & @20 \\
\midrule
\emph{Baselines} &&&&&&&&&\\
\midrule
BM25 & 20.9 & 53.8 & 62.7 & 20.9 & 53.8 & 62.7 & 20.9 & 53.8 & 62.7 \\
Dense Passage Retrieval & 25.8 & 62.6 & 70.9 & 32.8 & 74.9 & 82.6 & 35.5 & 78.7 & 86.1 \\
Cluster Integer IDs  & 38.4 & 64.2 & 69.4 & 40.2 & 67.5 & 72.7 & 40.8 & 68.2 & 73.0 \\
TSGen \citep{tsgen} & 28.8 & 67.1 & 73.6 & 29.2 & 67.6 & 74.4 & 30.3 & 71.8 & 78.3 \\
\midrule
\emph{Summarization-based IDs} &&&&&&&&&\\
\midrule
BM25 Top-30   & 36.5 & 66.1 & 70.9 & 36.8 & 66.1 & 71.1 & 37.0 & 68.2 & 72.8 \\
First 30 Tokens     & 41.9 & 66.0 & 69.9 & 43.3 & 67.6 & 71.6 & 47.7 & 71.2 & 74.4 \\
{\method}     & 39.2 & 69.2 & 74.0 & 40.5 & 70.7 & 75.2 & 40.9 & 74.9 & 80.2 \\
\midrule
\emph{Summarization-based IDs with Joint Decoding} &&&&&&&&&\\
\midrule
First 30 Tokens w/ Joint Dec & \textbf{49.1} & \textbf{78.7} & \textbf{82.6} & \textbf{49.7} & \textbf{79.2} & \textbf{83.1} & \textbf{55.3} & \textbf{83.0} & \textbf{86.4} \\
{\method} w/ Joint Dec   & 41.3 & 77.3 & 82.5 & 41.3 & 77.0 & 82.9 & 42.3 & 78.0 & 84.0 \\

\bottomrule
\end{tabular}}
\caption{Recall for Natural Questions. Recall refers to the percentage of queries in the evaluation set for which the ground-truth document ID was produced in the top-1, top-10, and top-20 candidates from constrained beam search decoding. NQ 1k, 10k, and 100k refer to the number of training query-document pairs used to finetune the LM. `Joint Dec' refers to joint decoding with the small, task-specific 160M parameter LM and a large 12B parameter LM with in-context learning.}
\label{tab:nq}
\end{table*}

%% file: figures/model_size.tex
\begin{figure*}[h]
\centering
\includegraphics[width=0.65\textwidth]{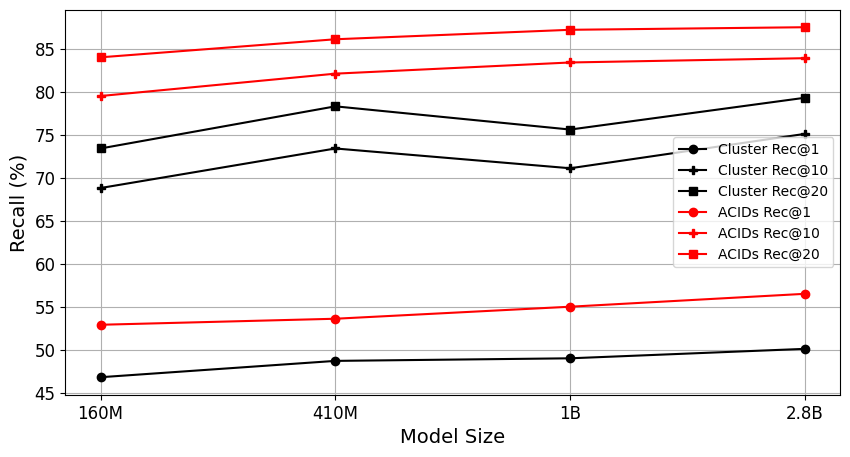}
\caption{Recall versus the number of parameters in the LM on the MSMARCO 100k dataset.}
\label{fig:model_size}
\end{figure*}

%% file: figures/beams.tex
\begin{table*}
\centering
\small
\begin{tabular}{rcccccc}

\toprule
 & \multicolumn{3}{c}{Cluster IDs} & \multicolumn{3}{c}{{\method}s} \\
 & Rec@1 & @10 & @20 & Rec@1 & @10 & @20 \\
\midrule
Beam width 1 & 47.6 & -- & -- & 54.0 & -- & -- \\
 2 & 48.7 & -- & -- & 56.0 & -- & -- \\
 4 & 49.0 & -- & -- & 55.7 & -- & -- \\
 8 & 48.8 & -- & -- & 56.5 & -- & -- \\
 16 & 49.0 & 71.0 & -- & 56.6 & 84.1 & -- \\
 20 & 49.0 & 71.1 & 75.6 & 55.0 & 83.4 & 87.2 \\
 30 & 49.0 & 70.9 & 75.6 & 56.4 & 84.1 & 88.3 \\
 40 & 49.0 & 70.9 & 75.6 & 56.5 & 84.1 & 88.3 \\
 50 & 49.0 & 70.9 & 75.6 & 56.5 & 84.1 & 88.4 \\
\bottomrule
\end{tabular}
\caption{Recall of the 1B-parameter model versus beam width on the MSMARCO 100k dataset.}
\label{fig:beams}
\end{table*}

%% file: tables/num_first_words.tex
\begin{table}[h]
\small
    \centering
    
    \begin{tabular}{@{}lcccccc@{}}
        \toprule
         & \multicolumn{3}{c}{MSMARCO 100k} & \multicolumn{3}{c}{NQ 100k} \\
        & Rec@1 & @10 & @20 & Rec@1 & @10 & @20 \\
        \midrule
        First 10 & 51.1 & 75.8 & 79.8 & 46.9 & 65.0 & 67.6 \\
        First 20 & 50.0 & 77.0 & 80.4 & 46.9 & 69.1 & 72.3 \\
        First 30 & 51.8 & 76.0 & 79.6 & 47.7 & 71.2 & 74.4 \\
        First 40 & 49.8 & 75.8 & 79.2 & 49.9 & 72.3 & 75.4 \\
        \bottomrule
    \end{tabular}
    \caption{Recall on MSMARCO and NQ 100k versus the length of the document ID. Here, we use the extractive summarization ID based on the first 10, 20, 30, or 40 tokens of each document.}
    \label{tab:num_first_words}
\end{table}

%% file: sections/related.tex
\section{Related Work}


\citet{neural_indexing} explore a number of techniques for creating document IDs for generative retrieval, including atomic document IDs, randomly assigned integer IDs, and semantic IDs based on hierarchical clustering. The last technique was found to be the most effective, where the document IDs with were formed via hierarchical $k$-means clustering on BERT-based document vectors. The main difference between that approach and ours is that, during finetuning, their approach requires learning the ``semantics'' of the cluster IDs, while ours uses natural language phrases that are already in some sense familiar to the pretrained model. \citet{pawa} also used IDs based on hierarchical clustering with BERT embeddings and proposed the prefix-aware weight-adaptor (PAWA) modification, where a separate decoder was trained to produce level-specific linear projections to modify the ID decoder's outputs at each timestep. The authors also incorporated synthetic queries from a doc2query model to augment the user-generated queries in the dataset. \citet{millions} scale the cluster ID-based approach to generative retrieval to millions of documents, and explore the impact of adding synthetic queries for documents that do not have a query sourced from a user.

The aforementioned papers used IDs that were not optimized for the retrieval task, but other work has explored creating document IDs in a retrieval-aware manner. In \citet{learning_to_tokenize}, the document IDs are treated as a sequence of fixed-length latent discrete variables which are learned via a document reconstruction loss and the generative retrieval loss. However, the authors reported that this method does experience collisions, as some documents are assigned to the same latent integer ID sequence, though  the collision rate was not reported.

\citet{ngram_retrieval} proposed a model that, given a query, generates the n-grams that should appear in the relevant documents. All documents that contain the generated n-grams are then retrieved and reranked to produce the final search results. (This is in contrast our approach, which seeks to associate a unique ID to each document for generative retrieval.) The authors propose several methods for reranking based on n-gram scores produced by the LM. However, the n-gram generation and reranking approach does not always outperform the dense retrieval baseline. \citet{tsgen} creates document IDs by selecting terms from the document based on relevance scores that are learned using a contrastive loss and BERT embeddings.

In addition, there is a substantial body of work that involves model-generated text and retrieval. \citet{entity_retrieval} generate the text representation of entities autoregressively instead of treating entities as atomic labels in a (potentially very large) vocabulary. \citet{doc2query} use an encoder-decoder model to generate synthetic queries for each document in the index and concatenate them together to improve retrieval performance. The expanded documents are indexed using Anserini and BM25. Synthetic queries from these `doc2query' models are also used for data augmentation in generative retrieval. \citet{query_expansion} use pretrained language models to expand queries with relevant contexts (e.g., appending the title of a relevant passage to the query, etc.) for retrieval and open-domain question answering. 

%% file: sections/discussion.tex
\section{Conclusion}

We have demonstrated that summarization-based document IDs are highly effective for generative retrieval. Our results show a clear improvement in retrieval performance on the Natural Questions and MSMARCO datasets versus both cluster-based integer IDs and other keyword-based document IDs. In direct comparisons, abstractive keyphrases work well versus other types of IDs. Surprisingly, we found that the first 30 tokens of a document also works very well among the IDs we tried, but we have not seen this fact documented in the generative retrieval literature. The choice of ID is clearly a major factor in retrieval performance, and we expect that future work will explore other possibilities for creating effective natural-language document IDs.

We also observed that the extractive summarization approach (i.e., first-30 tokens as ID) outperforms the abstractive \method approach for the long Wikipedia articles in the NQ dataset but not for the shorter snippets in the MSMARCO dataset. Clearly, the characteristics of the documents that are indexed affects generative retrieval, and in the case of Wikipedia documents, the initial sentences tend to be an overview of the rest of the article. As the field of generative retrieval continues to evolve, optimizing document ID generation for specific use cases and document collections may become an important area of study.





%% file: acl_latex.bbl
\begin{thebibliography}{19}
\providecommand{\natexlab}[1]{#1}

\bibitem[{Bajaj et~al.(2016)Bajaj, Campos, Craswell, Deng, Gao, Liu, Majumder, McNamara, Mitra, Nguyen et~al.}]{msmarco}
Payal Bajaj, Daniel Campos, Nick Craswell, Li~Deng, Jianfeng Gao, Xiaodong Liu, Rangan Majumder, Andrew McNamara, Bhaskar Mitra, Tri Nguyen, et~al. 2016.
\newblock \href {https://arxiv.org/abs/1611.09268} {{MS MARCO}: A human generated machine reading comprehension dataset}.
\newblock In \emph{Proc.\ of CoCo}.

\bibitem[{Bevilacqua et~al.(2022)Bevilacqua, Ottaviano, Lewis, Yih, Riedel, and Petroni}]{ngram_retrieval}
Michele Bevilacqua, Giuseppe Ottaviano, Patrick Lewis, Scott Yih, Sebastian Riedel, and Fabio Petroni. 2022.
\newblock Autoregressive search engines: Generating substrings as document identifiers.
\newblock \emph{Advances in Neural Information Processing Systems}, 35:31668--31683.

\bibitem[{Biderman et~al.(2023)Biderman, Schoelkopf, Anthony, Bradley, O'Brien, Hallahan, Khan, Purohit, Prashanth, Raff et~al.}]{pythia}
Stella Biderman, Hailey Schoelkopf, Quentin Anthony, Herbie Bradley, Kyle O'Brien, Eric Hallahan, Mohammad~Aflah Khan, Shivanshu Purohit, USVSN~Sai Prashanth, Edward Raff, et~al. 2023.
\newblock Pythia: A suite for analyzing large language models across training and scaling.
\newblock \emph{arXiv preprint arXiv:2304.01373}.

\bibitem[{De~Cao et~al.(2020)De~Cao, Izacard, Riedel, and Petroni}]{entity_retrieval}
Nicola De~Cao, Gautier Izacard, Sebastian Riedel, and Fabio Petroni. 2020.
\newblock Autoregressive entity retrieval.
\newblock \emph{arXiv preprint arXiv:2010.00904}.

\bibitem[{Devlin et~al.(2019)Devlin, Chang, Lee, and Toutanova}]{bert}
Jacob Devlin, Ming-Wei Chang, Kenton Lee, and Kristina Toutanova. 2019.
\newblock \href {https://arxiv.org/abs/810.04805} {{BERT}: Pre-training of deep bidirectional transformers for language understanding}.
\newblock In \emph{Proc.\ of NAACL}.

\bibitem[{Joshi et~al.(2017)Joshi, Choi, Weld, and Zettlemoyer}]{trivia_qa}
Mandar Joshi, Eunsol Choi, Daniel~S Weld, and Luke Zettlemoyer. 2017.
\newblock Triviaqa: A large scale distantly supervised challenge dataset for reading comprehension.
\newblock \emph{arXiv preprint arXiv:1705.03551}.

\bibitem[{Karpukhin et~al.(2020)Karpukhin, Oguz, Min, Lewis, Wu, Edunov, Chen, and Yih}]{dpr}
Vladimir Karpukhin, Barlas Oguz, Sewon Min, Patrick Lewis, Ledell Wu, Sergey Edunov, Danqi Chen, and Wen-tau Yih. 2020.
\newblock \href {https://arxiv.org/abs/2004.04906} {Dense passage retrieval for open-domain question answering}.
\newblock In \emph{Proc.\ of EMNLP}.

\bibitem[{Kwiatkowski et~al.(2019)Kwiatkowski, Palomaki, Redfield, Collins, Parikh, Alberti, Epstein, Polosukhin, Devlin, Lee et~al.}]{nq}
Tom Kwiatkowski, Jennimaria Palomaki, Olivia Redfield, Michael Collins, Ankur Parikh, Chris Alberti, Danielle Epstein, Illia Polosukhin, Jacob Devlin, Kenton Lee, et~al. 2019.
\newblock Natural questions: a benchmark for question answering research.
\newblock \emph{Transactions of the Association for Computational Linguistics}, 7:453--466.

\bibitem[{Loshchilov and Hutter(2017)}]{adamW}
Ilya Loshchilov and Frank Hutter. 2017.
\newblock Decoupled weight decay regularization.
\newblock \emph{arXiv preprint arXiv:1711.05101}.

\bibitem[{Mao et~al.(2020)Mao, He, Liu, Shen, Gao, Han, and Chen}]{query_expansion}
Yuning Mao, Pengcheng He, Xiaodong Liu, Yelong Shen, Jianfeng Gao, Jiawei Han, and Weizhu Chen. 2020.
\newblock Generation-augmented retrieval for open-domain question answering.
\newblock \emph{arXiv preprint arXiv:2009.08553}.

\bibitem[{Nogueira et~al.(2019)Nogueira, Yang, Lin, and Cho}]{doc2query}
Rodrigo Nogueira, Wei Yang, Jimmy Lin, and Kyunghyun Cho. 2019.
\newblock Document expansion by query prediction.
\newblock \emph{arXiv preprint arXiv:1904.08375}.

\bibitem[{Pradeep et~al.(2023)Pradeep, Hui, Gupta, Lelkes, Zhuang, Lin, Metzler, and Tran}]{millions}
Ronak Pradeep, Kai Hui, Jai Gupta, Adam~D Lelkes, Honglei Zhuang, Jimmy Lin, Donald Metzler, and Vinh~Q Tran. 2023.
\newblock How does generative retrieval scale to millions of passages?
\newblock \emph{arXiv preprint arXiv:2305.11841}.

\bibitem[{Raffel et~al.(2020)Raffel, Shazeer, Roberts, Lee, Narang, Matena, Zhou, Li, and Liu}]{t5}
Colin Raffel, Noam Shazeer, Adam Roberts, Katherine Lee, Sharan Narang, Michael Matena, Yanqi Zhou, Wei Li, and Peter~J. Liu. 2020.
\newblock \href {http://jmlr.org/papers/v21/20-074.html} {Exploring the limits of transfer learning with a unified text-to-text transformer}.
\newblock \emph{JMLR}.

\bibitem[{Sun et~al.(2024)Sun, Yan, Chen, Wang, Zhu, Ren, Chen, Yin, Rijke, and Ren}]{learning_to_tokenize}
Weiwei Sun, Lingyong Yan, Zheng Chen, Shuaiqiang Wang, Haichao Zhu, Pengjie Ren, Zhumin Chen, Dawei Yin, Maarten Rijke, and Zhaochun Ren. 2024.
\newblock Learning to tokenize for generative retrieval.
\newblock \emph{Advances in Neural Information Processing Systems}, 36.

\bibitem[{Tay et~al.(2022)Tay, Tran, Dehghani, Ni, Bahri, Mehta, Qin, Hui, Zhao, Gupta et~al.}]{neural_indexing}
Yi~Tay, Vinh Tran, Mostafa Dehghani, Jianmo Ni, Dara Bahri, Harsh Mehta, Zhen Qin, Kai Hui, Zhe Zhao, Jai Gupta, et~al. 2022.
\newblock Transformer memory as a differentiable search index.
\newblock \emph{Advances in Neural Information Processing Systems}, 35:21831--21843.

\bibitem[{Wang et~al.(2022)Wang, Hou, Wang, Miao, Wu, Chen, Xia, Chi, Zhao, Liu et~al.}]{pawa}
Yujing Wang, Yingyan Hou, Haonan Wang, Ziming Miao, Shibin Wu, Qi~Chen, Yuqing Xia, Chengmin Chi, Guoshuai Zhao, Zheng Liu, et~al. 2022.
\newblock A neural corpus indexer for document retrieval.
\newblock \emph{Advances in Neural Information Processing Systems}, 35:25600--25614.

\bibitem[{Yang et~al.(2017)Yang, Fang, and Lin}]{anserini}
Peilin Yang, Hui Fang, and Jimmy Lin. 2017.
\newblock Anserini: Enabling the use of lucene for information retrieval research.
\newblock In \emph{Proceedings of the 40th international ACM SIGIR conference on research and development in information retrieval}, pages 1253--1256.

\bibitem[{Zhang et~al.(2024)Zhang, Liu, Zhou, Dou, Liu, and Cao}]{tsgen}
Peitian Zhang, Zheng Liu, Yujia Zhou, Zhicheng Dou, Fangchao Liu, and Zhao Cao. 2024.
\newblock Generative retrieval via term set generation.
\newblock \emph{arXiv preprint arXiv:2305.13859}.

\bibitem[{Zhuang et~al.(2023)Zhuang, Ren, Shou, Pei, Gong, Zuccon, and Jiang}]{dsi_qg}
Shengyao Zhuang, Houxing Ren, Linjun Shou, Jian Pei, Ming Gong, Guido Zuccon, and Daxin Jiang. 2023.
\newblock Bridging the gap between indexing and retrieval for differentiable search index with query generation.
\newblock \emph{The First Workshop on Generative Information Retrieval at SIGIR}.

\end{thebibliography}
